\pdfoutput=1
\documentclass[10pt,twocolumn,letterpaper]{article}

\usepackage{cvpr}              

\usepackage{booktabs}
\usepackage{times}
\usepackage{epsfig}
\usepackage{graphicx}
\usepackage{amsmath}
\usepackage{caption}
\usepackage{enumitem}
\usepackage{amssymb}
\usepackage{tabularx}
\usepackage{multirow}
\usepackage{ctable}

\usepackage[pagebackref,breaklinks,colorlinks]{hyperref}

\usepackage[capitalize]{cleveref}
\crefname{section}{Sec.}{Secs.}
\Crefname{section}{Section}{Sections}
\Crefname{table}{Table}{Tables}
\crefname{table}{Tab.}{Tabs.}

\def\cvprPaperID{*****} 
\def\confName{CVPR}
\def\confYear{2023}

\begin{document}

\title{You Only Train Once: Multi-Identity Free-Viewpoint\\Neural Human Rendering from Monocular Videos}

\author{Jaehyeok Kim$^1$ \qquad Dongyoon Wee$^2$ \qquad Dan Xu$^1$\\
$^1$The Hong Kong University of Science and Technology \quad $^2$Clova AI, NAVER Corp.\\
{\tt\small jkimbf@connect.ust.hk, dongyoon.wee@navercorp.com, danxu@cse.ust.hk}
}
\twocolumn[{%
\renewcommand\twocolumn[1][]{#1}%
\maketitle 
\begin{center}
    \centering
    \captionsetup{type=figure}
    \vspace{-15pt}
    \includegraphics[width=1.0\textwidth]{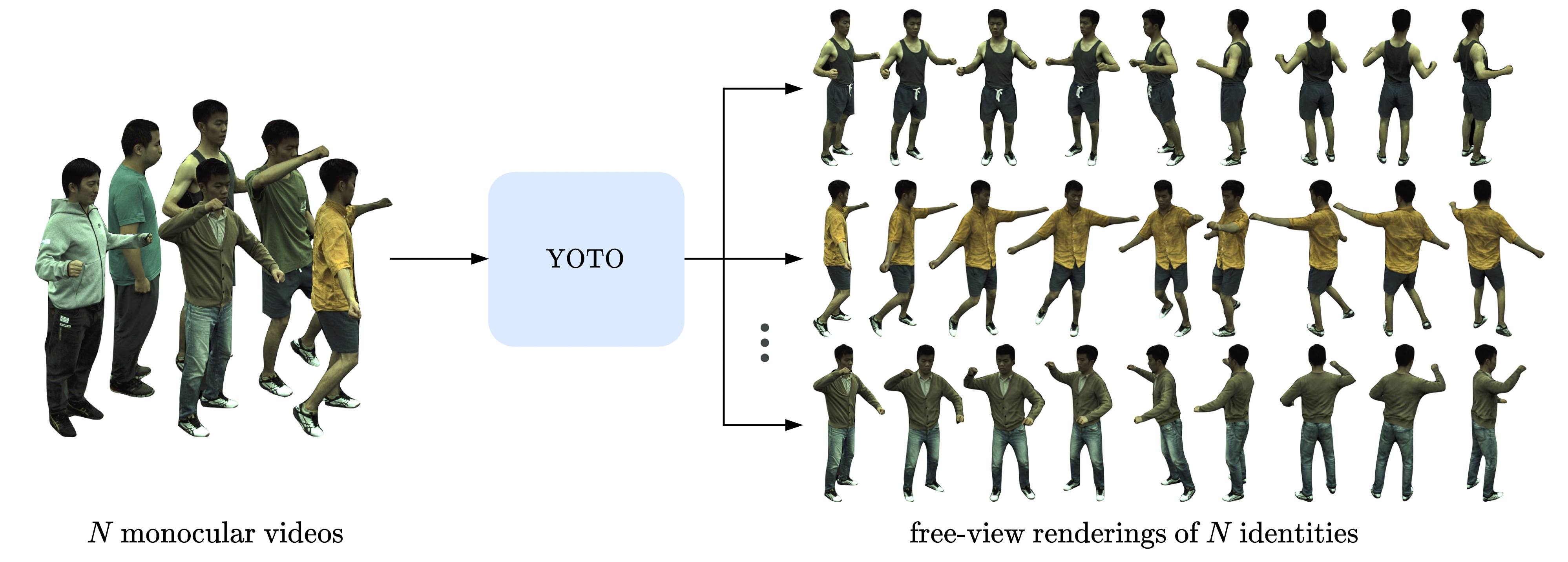} 
    \vspace{-20pt}
    \label{fig:teaser}
    \captionof{figure}{The proposed YOTO is a dynamic human generation framework, which can simultaneously train and model multiple people with distinct appearances and motions, while also allowing coherent rendering of them in the unified framework in high fidelity. The identity animation can be performed by using either seen poses or novel poses.}
\end{center}
}]
\maketitle

\setlength{\abovedisplayskip}{2.5pt}
\setlength{\belowdisplayskip}{2.5pt}
\begin{abstract}
We introduce You Only Train Once (YOTO), a dynamic human generation framework, which performs free-viewpoint rendering of different human identities with distinct motions, via only one-time training from monocular videos. Most prior works for the task require individualized optimization for each input video that contains a distinct human identity, leading to a significant amount of time and resources for the deployment, thereby impeding the scalability and the overall application potential of the system. In this paper, we tackle this problem by proposing a set of learnable identity codes to expand the capability of the framework for multi-identity free-viewpoint rendering, and an effective pose-conditioned code query mechanism to finely model the pose-dependent non-rigid motions. YOTO optimizes neural radiance fields (NeRF) by utilizing designed identity codes to condition the model for learning various canonical T-pose appearances in a single shared volumetric representation. Besides, our joint learning of multiple identities within a unified model incidentally enables flexible motion transfer in high-quality photo-realistic renderings for all learned appearances. This capability expands its potential use in important applications, including Virtual Reality. We present extensive experimental results on ZJU-MoCap and PeopleSnapshot to clearly demonstrate the effectiveness of our proposed model. YOTO shows state-of-the-art performance on all evaluation metrics while showing significant benefits in training and inference efficiency as well as rendering quality. The code and model will be made publicly available soon.
\end{abstract}

\section{Introduction}
Novel view synthesis of a person with dynamic motions from a monocular video is an especially challenging and long-standing problem. Unlike other similar tasks dealing with dynamic scenes, it requires modeling not only complicated motions generated by body joints but also non-rigidities of finer-granularity components such as the body and clothes. Moreover, the monocular setting further complicates the problem as information about every body motion from a single-image view is extremely limited. Therefore, the free-viewpoint rendering of moving people has mostly been investigated under multi-view settings without taking into account non-rigid fine-grained motions.

\par Recently, Weng~\etal~\cite{weng2022humannerf} successfully address the above-mentioned problem with an architecture comprising structures for learning of a motion field and a NeRF~\cite{mildenhall2020nerf}.
Although they address the monocular free-viewpoint rendering with state-of-the-art performance, its identity-specific nature largely limits its potential for practical application scenarios. It requires an independent model that is trained from scratch for every human identity from each monocular video. This constraint severely downgrades the efficiency and generalization performance of the method. For instance, it is clearly not scalable if there is a considerable number of human-specific videos to be learned, as the overall model size and the training time would significantly increase, proportional to the number of identities following their pipeline. Besides, it is obviously not flexible to perform any interactions between the different identities (\textit{e.g.}, performing motion transfer among the identities in different videos), due to the fact that there are no correlations constructed during training and inference among the different human models. We believe that each video comprises generic and distinctive information and the generic one could be learned collaboratively from different videos by having an unified NeRF framework, thus benefitting the rendering performance as well.

\par To target the issues above-discussed, in this paper, we propose YOTO, a more versatile framework for the monocular free-viewpoint rendering of people with distinct motions. We introduce an effective set of learnable identity codes into the framework to enable learning {global} human-specific representations, which can be utilized in our framework as a perfect switcher to allow multi-subject modeling and renderings, using a single unified model by only \emph{one-time} optimization.
Furthermore, in this paper, we present a novel mechanism for querying a separate identity code to learn identity-specific non-rigid motions. This involves utilizing cross-attention between the identity code and the 3D pose of the current frame. 
By adopting this process, our framework is able to extract discriminative features that are tailored to each identity with a particular pose, resulting in better non-rigid motion estimation.
YOTO conditions two different NeRFs, each for non-rigid motions and appearances learned in a unified manner. Unlike Weng~\etal~\cite{weng2022humannerf}, it incorporates all subjects in interest at the same time for training while not requiring proportionally longer training time.
This significantly improves the efficiency of the framework on a number of people and further enhances both qualitative and quantitative performance.
Overall, YOTO achieves state-of-the-art performance on free-viewpoint rendering of multiple moving people while showing remarkable enhancements in flexibility and training/inference efficiency compared to~\cite{weng2022humannerf}.
We present our experimental results on ZJU-MoCap~\cite{peng2021neuralbody} and PeopleSnapshot~\cite{alldieck2018peoplesnapshot} to demonstrate that YOTO can competently handle hard cases (\emph{e.g.}~input videos in the wild) and achieve state-of-the-art performances.
\par In summary, the contribution of our work is threefold:
\begin{itemize}[leftmargin=0.9em, topsep=0.05em]
    \setlength\itemsep{-0.25em}
    \item We resolve the issue of subject-specific training by proposing a new framework with learnable identity codes that allows multi-human-identity representation learning.
    \item The proposed framework, YOTO, better models pose-dependent non-rigid motions by conditioning the non-rigid modeling with a pose-conditioned code queried by cross-attention.
    \item YOTO not only achieves state-of-the-art performance on all quantitative metrics but also remarkably improves the model efficiency. Incidentally, YOTO can animate all the learned appearances of different identities in high fidelity with any novel poses, thereby enabling high-quality animations for various applications.
\end{itemize}

\begin{figure*}[!t]
\centering
  \includegraphics[width=1\linewidth]{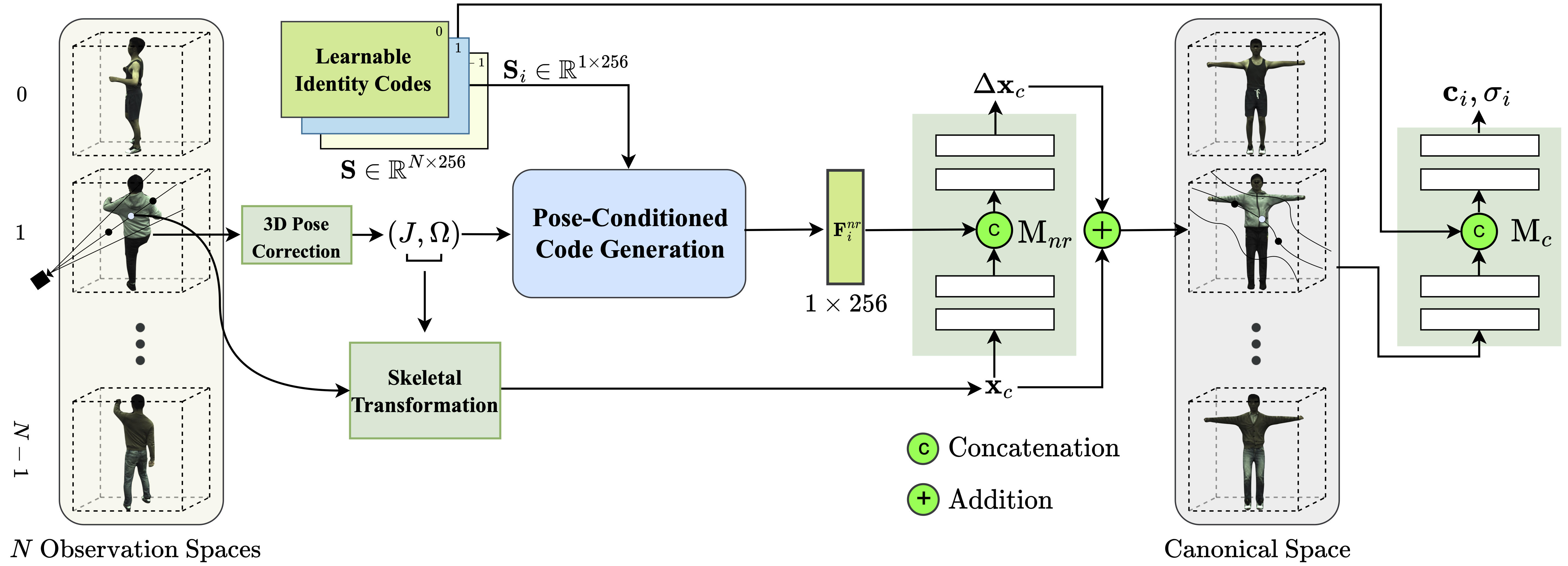}%
  \vspace{-7pt}
  \caption{Overview of the proposed framework YOTO for free-view human rendering from monocular videos. Our framework is able to simultaneously train multiple identities while also achieving state-of-the-art performances on rendering quality. This is achieved by the proposed learnable identity codes and the body-pose-conditioned code generation module for subject-specific non-rigid motion estimation and canonical radiance field prediction.}
  \vspace{-5pt}
  \label{fig:overview}
\end{figure*}
\section{Related Work}
We review closely related works from three perspectives, \emph{i.e.}, deformable neural rendering, neural rendering of humans, and monocular neural human rendering. 
\par\noindent\textbf{Deformable neural rendering.} NeRF~\cite{mildenhall2020nerf} leverages a multi-layer perceptron (MLP) to learn a static 3D representation of a scene from a dense set of images from diverse viewpoints. Among various research directions, recent works have enhanced NeRF in terms of its efficiency and performance.
For instance, several works~\cite{garbin2021fastnerf, reiser2021kilonerf, yu2021plenoctrees, hedman2021snerg, yu_and_fridovichkeil2021plenoxels, mueller2022instant, chen2022tensorf} boost the efficiency of NeRF in training or inference stages to allow more practical usage.
Some others~\cite{yu2020pixelnerf, jain2021dietnerf, chen2021mvsnerf, wang2021ibrnet, liu2022neuray} extend NeRF to handle the sparse view setting, which adapts NeRF to more realistic training scenarios and expands its practicality.
However, it has been observed that they are restricted to static scenes while the majority of the real-world objects are dynamic.

Recent works including ~\cite{gao2021dynerf, li2020neuralflow, pumarola2020dnerf, park2021nerfies, tretschk2021nonrigid, park2021hypernerf, xian2021spacetime} have broadened the modeling capabilities of NeRF to dynamic scenes containing movements or deformations. Park~\etal~\cite{park2021nerfies, park2021hypernerf} handle natural deformations on faces by introducing an MLP to estimate per-point deformations and~\cite{park2021hypernerf} additionally models topological changes by proposing a canonical hyper-space.
Pumarola~\etal~\cite{pumarola2020dnerf} and Tretschk~\etal~\cite{tretschk2021nonrigid} propose the same idea as~\cite{park2021nerfies} to model dynamic objects and non-rigidity, respectively.
It should be noted that these motions and deformations are small and simple; otherwise, the aforementioned approaches would not exhibit the anticipated level of performance.
Nonetheless, there are deformations in the world spanning from small-scale motions to more complex articulated motions. In contrast, our approach enables learning of human-articulated motions from monocular videos, and performing free-viewpoint rendering of a human performer at any time frame of the videos.

\par\noindent
\textbf{Neural rendering of humans.} 
As our approach targets the problem of neural human rendering, we discuss related works in this direction.
Martin-Brualla~\etal~\cite{martin-brualla2018lookinggood} propose a neural re-rendering approach via U-Net-like architecture for reducing the generated artifacts. By utilizing a few calibration images of the target subject, Pandey~\etal~\cite{pandey2019semiparam} introduce semi-parametric learning from a single or few input RGBD frames.
Similarly, Liu~\etal~\cite{liu2020NeuralHumanRendering} propose to use a character model to generate priors for learning time-coherent dynamic textures.
Wu~\etal~\cite{wu2020multi} learn explicit 3D features on point clouds produced from multi-view stereo~\cite{johannes2016mvs} and use U-Net for free-viewpoint rendering. To enable learning implicit representations from a highly sparse set of input views, structured latent codes are introduced by Peng~\etal~\cite{peng2021neural} to be applied on a shared deformable mesh (SMPL~\cite{loper2015smpl}).
There are also several prior works exploring learning animatable avatars ~\cite{jiang2022neuman, jiang2022selfrecon, wang2022arah, huang2022elicit}, which however are based on explicit human parametric models~\cite{loper2015smpl}, instead of implicit representations. Besides, Jiang~\etal~\cite{jiang2022instantavatar} adopt Instant-NGP~\cite{mueller2022instant} to boost the training efficiency. While most of these existing works either use the explicit SMPL as a prior or require multi-view videos, our approach does not rely on the parametric models and utilizes only monocular videos and 3D poses as inputs.

\par \noindent
\textbf{Monocular neural human rendering.} 
Recently, Weng~\etal~\cite{weng2022humannerf} propose an approach to conduct free-viewpoint rendering of a human performer within a monocular video.
It learns a canonical T-pose representation of the performer by modeling rigid body motions and pose-dependent non-rigid motions.
However, it has a severe limitation of subject-specific modeling, which requires a new model to be trained from scratch for \textit{each} input monocular video. 
In this paper, we introduce a set of novel learnable identity codes and an effective pose-conditioned code query mechanism, which allows our single model to represent an arbitrary number of human subjects while outperforming the baseline~\cite{weng2022humannerf} on all evaluation metrics.

\section{The Proposed YOTO Approach}
\subsection{Framework Overview}
Given a number of monocular videos each containing a single distinct human subject, the proposed YOTO framework learns discriminative representations of all the moving identities by one-time training, for free-viewpoint rendering, as shown in the framework overview (see Fig.~\ref{fig:overview}). Specifically, we propose a novel idea to enable a simultaneous optimization of a collaborative canonical representation of all the subjects, via introducing a set of \emph{learnable} identity codes.
To condition the learning of identity-specific non-rigid motions, we further propose a module to generate pose-conditioned identity codes. 

For each identity in a monocular video, the module utilizes a learnable identity code and joint poses corresponding to the target subject as input, and produces a subject-specific pose-conditioned code by a cross-attention mechanism. The generated code is then embedded into an MLP to condition the non-rigid motion estimation. It accepts an input point from a skeletal transformation~\cite{weng2022humannerf} that maps input deformed joint poses to a canonical T-pose based on subject-specific blend weights with inverse linear-blend skinning. Then, for the transformed canonical points with non-rigid motions applied, we further enable multi-identity rendering by conditioning a canonical MLP with the proposed learnable identity codes, to versatilely predicts the radiance and density that correspond to each different target subject.

\subsection{Preliminaries: HumanNeRF}
In this section, we describe two technical components of HumanNeRF that our proposed framework is based on, \ie,~the skeletal motion estimation which learns sets of blend weights for skeletal poses, and the pose correction which corrects estimated error-prone 3D body poses.

\begin{figure}[!t]
\begin{center}
   \includegraphics[width=\linewidth]{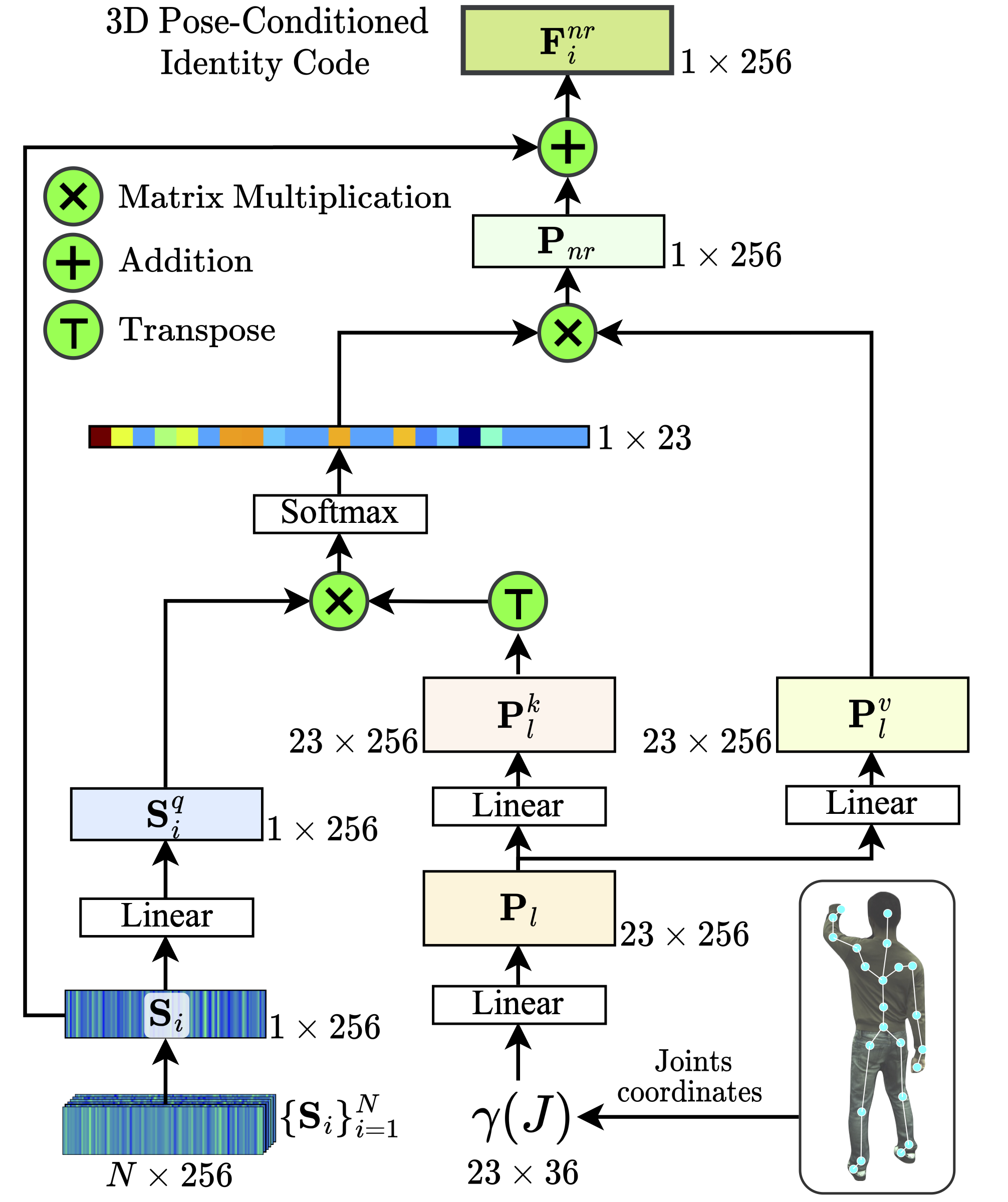}
   \vspace{-2.5em}
\end{center}
    \vspace{-5pt}
    \caption{Illustration of the proposed pose-conditioned identity code generation. It accepts a learnable identity code and joint poses of a subject as input and produces an identity-specific code for non-rigid motion estimation.}
    \vspace{-13pt}
\label{fig:attention}
\end{figure}

\par\noindent
\textbf{Skeletal motion.}
The skeletal motion learns an inverse formulation of the linear blend skinning, which is an algorithm to render high-order deformation of objects caused by low-order skeletons~\cite{jacobson2014skinning}.
It transforms a vertex from the canonical pose to the target pose by weighted-summing a point with a set of skinning weights that describe the degree of influence of each bone and transformation matrices.
Weng~\etal~\cite{weng2022humannerf} reformulate the linear blend skinning to transform a point $\mathbf{x}$ in the observation space to a point $\mathbf{x}_c$ in the canonical space, which can be written as:
\begin{gather}
    \textbf{x}_c = \displaystyle\sum_{k=1}^K w^o_{k}(R_k\textbf{x} + \textbf{t}_k),
    w^o_{k} = \frac{w^c_k(R_k\textbf{x} + \textbf{t}_k)}{\textstyle\sum_{j=1}^K w^c_{j}(R_j\textbf{x} + \textbf{t}_j)}, \label{eq:1}
\end{gather}
where $w^o_{k}$ and $w^c_{k}$ respectively represent observation and canonical skinning weights of bone $k$ on a point $\textbf{x}$ in the observation space, and $R_k$ and $\textbf{t}_k$ are the rotation and translation of the bone $k$.
Given Eq.~\ref{eq:1}, HumanNeRF optimizes a CNN network to predict the set of canonical skinning weights $\{w_k\}^K_{k=1}$ for each observation point where $K$ is the total number of joints.
YOTO adopts the same formulations for skeletal motion by having distinct motion priors of each identity taken by the CNN network as inputs.

\par\noindent\textbf{Pose correction.}
As pointed out by Weng~\etal~\cite{weng2022humannerf}, the aforementioned input 3D body pose $(J, \Omega)$ ($J$ and $\Omega$ denote the joint locations and orientations, respectively) tends to be error-prone as it is from an off-the-shelf pose estimator.
Therefore, it introduces a pose correction MLP that takes the joint orientations $\Omega = \{\Omega_k\}^K_{k=1}$ and predicts their offsets for pose refinement. 
Over multiple training iterations, the joint angles of the 3D body pose undergo continuous refinements, thus resulting in state-of-the-art performance.

\subsection{Learnable Identity Codes}
We now introduce the proposed learnable identity codes that are essential and effective for enabling discriminative multi-identity rendering using only one-time training. 
We propose to impose the learnable identity codes to condition both the non-rigid MLP $M_{nr}$ for fine-grained motion estimation and the canonical MLP $M_c$ for identity-specific radiance field prediction. To achieve this goal, a set of $N$ {learnable} identity codes $\{\mathbf{S}_i\in\mathbb{R}^{256}\}^N_{i=1}$ are defined corresponding to $N$ identities in the training video data, and each identity code $\mathbf{S}_i$ is learned globally based on gradient descent to represent each identity.  
$\mathbf{S}_i$ is jointly optimized with the objectives of the framework to obtain subject-specific representations. For each monocular video with a specific identity, we use its corresponding identity code as input, and thus the learned identity codes are discriminative. Then, we directly use the identity codes to condition the canonical MLP to predict the subject-specific radiance field, while for the non-rigid motion estimation, we further introduce an effective mechanism to generate pose-conditioned identity code to better facilitate the subject-specific fine-grained motion estimation. 

\subsection{3D Body Pose Conditioned Identity Code}
Each learned identity code can represent a subject globally from all the monocular videos. However, different video frames of the same identity may present distinct body motions. To further enhance the $N$ learnable identity codes $\{\mathbf{S}_i\}_{i=1}^{N}$ for the learning of non-rigid motions, we propose to employ the 3D body pose as guidance for generating a pose-condition identity code. A cross-attention mechanism is designed to perform interactions between the learnable identity codes and the 23 joint positions (\ie~$J = \{j_i\}^{23}_{i=1}$). A detailed overview of the mechanism is illustrated in Fig.~\ref{fig:attention}.
To learn implicit representations of the 3D body joints, we conduct positional encoding~~\cite{mildenhall2020nerf} for the input joint points. We first project each joint position into a higher dimension by using a sinusoidal positional encoding function, $\gamma(j_i) = \left(\sin(2^0 \pi j_i), \cos(2^0 \pi j_i), ..., \sin(2^{L-1} \pi j_i), \cos(2^{L-1} \pi j_i)\right)$, 
where $L$ is the number of frequency bands; $\gamma(\cdot)$ is independently applied to each joint. After this procedure, we generate a $36$-dimension representation for each input joint. We then project the encoded points (\ie, $\gamma(J)\in \mathbb{R}^{23\times 36}$) to an implicit pose code $\mathbf{P}_l$ by feeding it to a single linear layer with parameters $\mathbf{W}_p$ as below:
\begin{equation}
    \mathbf{P}_l = \mathbf{W}_p \cdot \gamma(j_i). \label{eq:3}
\end{equation}
Then, we generate one query signal $\mathbf{S}_i^q$ from the corresponding identity code $\mathbf{S}_i$ of the target identity via a projection matrix $\mathbf{W}_Q$ , and key and value signals from the pose code $\mathbf{P}_l$ via two other projection matrices $\mathbf{W}_K$ and $\mathbf{W}_V$ as:
\setlength{\abovedisplayskip}{3.0pt}
\setlength{\belowdisplayskip}{3.0pt}
\begin{equation}
    \mathbf{S}^q_i = \mathbf{W}_Q \cdot \mathbf{S}_i,\quad 
    \mathbf{P}^k_l = \mathbf{W}_K \cdot \mathbf{P}_l,\quad 
    \mathbf{P}^v_l = \mathbf{W}_V \cdot \mathbf{P}_l. \label{eq:4}
\end{equation}
\noindent
Finally, we generate the pose-conditioned identity code $\mathbf{F}_i^{nr}$ for the non-rigid motion estimation as:
\begin{equation}
    \mathbf{P}_{nr} = \text{softmax}(\mathbf{S}^q_i \cdot (\mathbf{P}^k_l)^\top) \cdot \mathbf{P}^v_l, \label{eq:5} \ \
    \mathbf{F}_i^{nr} = \mathbf{P}_{nr} + \mathbf{S}_i.
\end{equation}
We further employ this code for conditioning the learning of the non-rigid MLP (\ie $\mathbf{M}_{nr}$). This step is intended to encourage this MLP to estimate pose-coherent non-rigid motions. The details are discussed in Section \ref{sec:3.5}.
\setlength{\abovedisplayskip}{2.5pt}
\setlength{\belowdisplayskip}{2.5pt}

\begin{table*}[!t]
\centering
  \resizebox{1\linewidth}{!}{
    \begin{tabular}{ l | c c c | c c c | c c c } \toprule
        \multirow{2}{*}{} & \multicolumn{3}{c|}{Subject 377} & \multicolumn{3}{c|}{Subject 386} & \multicolumn{3}{c}{Subject 387} \\
        \cmidrule{2-10}
        & PSNR $\uparrow$ & SSIM $\uparrow$ & LPIPS$^*$ $\downarrow$ & PSNR $\uparrow$ & SSIM $\uparrow$ & LPIPS$^*$ $\downarrow$ & PSNR $\uparrow$ & SSIM $\uparrow$ & LPIPS$^*$ $\downarrow$ \\ 
        \midrule
        HumanNeRF~\cite{weng2022humannerf} & 
        30.39 & 0.9624 & 25.27 & 
        33.18 & 0.9629 & 30.29 & 
        28.11 & 0.9515 & 36.98 \\
        YOTO (Ours) & 
        \textbf{30.57} & \textbf{0.9698} & \textbf{21.88} & 
        \textbf{33.43} & \textbf{0.9655} & \textbf{26.11} & 
        \textbf{28.39} & \textbf{0.9534} & \textbf{34.55} \\ 
        \bottomrule
    \end{tabular}
    }
    \newline
    \vspace*{1 pt}
    \newline
  \resizebox{1\linewidth}{!}{
    \begin{tabular}{ l | c c c | c c c | c c c }
        \toprule
        \multirow{2}{*}{} & \multicolumn{3}{c|}{Subject 392} & \multicolumn{3}{c|}{Subject 393} & \multicolumn{3}{c}{Subject 394} \\
        \cmidrule{2-10}
        & PSNR $\uparrow$ & SSIM $\uparrow$ & LPIPS$^*$ $\downarrow$ & PSNR $\uparrow$ & SSIM $\uparrow$ & LPIPS$^*$ $\downarrow$ & PSNR $\uparrow$ & SSIM $\uparrow$ & LPIPS$^*$ $\downarrow$ \\ 
        \midrule
        HumanNeRF~\cite{weng2022humannerf} & 
        31.03 & 0.9580 & 33.99 & 
        28.29 & 0.9476 & 39.22 & 
        30.31 & 0.9507 & 34.64\\
        YOTO (Ours) & 
        \textbf{31.21} & \textbf{0.9598} & \textbf{31.06} & 
        \textbf{28.70} & \textbf{0.9504} & \textbf{35.63} & 
        \textbf{30.80} & \textbf{0.9535} & \textbf{32.11}\\ 
        \bottomrule
    \end{tabular}
    }
    \vspace{-5pt}
    \caption{Quantitative results on ZJU-MoCap dataset where LPIPS$^*$ $=$ LPIPS $\times 10^3$ following Weng~\etal~\cite{weng2022humannerf}. As there is no publicly available evaluation protocol from HumanNeRF (\eg~their used testing frames), we directly use their released checkpoints that achieved the performance mentioned in~\cite{weng2022humannerf} to evaluate on our evaluation protocol, in which we choose to evaluate on all existing testing frames instead of sampled frames, in order to have thorough and strict evaluations. YOTO outperforms the baseline on all metrics: PSNR, LPIPS$^*$, and SSIM. It should be noted that our model one-time trains all the identities, while HumanNeRF optimizes each identity separately.}
    \label{tab:quantitative}
    \vspace{-7pt}
\end{table*}

\subsection{Pose-Conditioned IDs for Non-rigid Motions}\label{sec:3.5}
~Learning the non-rigid body motions from the input monocular video is critical for generating natural and high-fidelity rendering results.
Therefore, we make use of an MLP to learn the point-specific non-rigid movements by predicting the corresponding offsets $\Delta\textbf{x}_c$ to the canonical point $\textbf{x}_c$.
We first apply the aforementioned positional encoding $\gamma(\cdot)$ to enable the learning of high-frequency details. We concatenate the relative joint coordinates $J$ to $\gamma(\textbf{x}_c)$ as an input to $\mathbf{M}_{nr}$ so that it can learn the pose-dependent non-rigid motions.
Moreover, we additionally concatenate the pose-conditioned identity code $\mathbf{F}^{nr}_i$ to the intermediate logits of $\mathbf{M}_{nr}$, thus allowing $\mathbf{M}_{nr}$ to render subject and pose-dependent non-rigid deformations as well.
We intend the code to play the role of assisting $\mathbf{M}_{nr}$ to learn the non-rigid motions, and thus we concatenate it with the representation of point $\mathbf{x}_c$ in the middle, which can be formulated as:
\begin{equation}
    \Delta\textbf{x}_c = \mathbf{M}_{nr}(J \oplus \gamma(\textbf{x}_c); \mathbf{F}^{nr}_i), \ \
    \textbf{x}_c = \textbf{x}_c + \Delta\textbf{x}_c. \label{eq:7}
\end{equation}
We simply add the estimated motion offsets from the pose condition to the input 3D canonical point $\textbf{x}_c$. As the result, the 3D canonical points reflect both the non-rigid motions and the subject-dependent 3D shape deformations.

\subsection{ID-Conditioned Canonical Representations}
To supervise the model training and render novel images, we now regress the radiance for each canonical point using another MLP.
We embed the learnable identity code $\mathbf{S}_i$ of the $i-$th identity in the training video, into the middle layer of the MLP via concatenating with the input point representation, so that the same canonical MLP $\mathbf{M}_c$ can also learn different independent subject-specific radiance fields. Benefiting from the identity codes as conditions, our YOTO framework optimizes only $1$ canonical space while representing $N$ different subjects.
The color $\mathbf{c}_i$ and density $\mathbf{\sigma}_i$ of each point can then be predicted by:
\begin{equation}
    \textbf{c}_i, \sigma_i = \mathbf{M}_c(\gamma(\textbf{x}_c); \ \mathbf{S}_{i}). \label{eq:8}
\end{equation}

\noindent
\textbf{Volume rendering.}
Following Mildenhall~\etal~\cite{mildenhall2020nerf}, we adopt stratified sampling and volume rendering to compute the estimated color of each ray.
We sample $M$ different points for each ray $\textbf{r}$ and integrate the colors and densities of them for subject $i$ as follows:
\begin{gather}
    C_i(\textbf{r}) = \displaystyle\sum_{m=1}^M T_m(1-\text{exp}(-\sigma_{i,m}\delta_{i,m}))\textbf{c}_{i,m}, \label{eq:9}
\end{gather}
\noindent
where $\delta_{i,m}$ is the adjacent distance from $m$-th to $m+1$-th sample and $T_m=\text{exp}(-\textstyle\sum_{n=1}^{m-1} \sigma_{i,n} \delta_{i,n})$.

\subsection{Optimization}
We perform one-time optimization of the model by training it with combined image frames of all $N$ subjects $\{I^i_1, I^i_2, ..., I^i_{F_i}\}^N_{i=1}$, where $F_i$ is the number of training frame for subject $i$.
For each training iteration, YOTO randomly selects one frame and samples rays regardless of the subject identities.
For fair comparisons to the baseline, we follow the setup of Weng~\etal~\cite{weng2022humannerf} as described below.
Our framework also samples rays in a patch $\hat{P_{F_i}}$ from an image $I^i_{F_i}$ to utilize the LPIPS~\cite{zhang2018perceptual} loss term.
The LPIPS loss term measures the perceptual distance between two image patches, and thus we also adopt it for more perceptually good renderings.
Our framework takes the features of patches extracted from the pre-trained VGGNet~\cite{simonyan2015vggnet} and computes the LPIPS loss term $\mathcal{L}_{\text{LPIPS}}$.
Moreover, we also compute an $L2$ loss term (\ie~$\mathcal{L}_2$) of the rendered RGB for each ray.
We combine the two loss terms with a coefficient $\lambda$ and write the overall optimization loss $\mathcal{L}_{o}$ for the whole framework as:
\begin{equation}
\begin{split}
    &\mathcal{L}_o = \mathcal{L}_\text{LPIPS} + \lambda \mathcal{L}_2, \hspace{7pt}
    \mathcal{L}_2 = \sum (C_i(\textbf{r}) - C_i^{GT}(\textbf{r}))^2, \\
    &\mathcal{L}_\text{LPIPS} = \text{LPIPS}(VGG(\hat{P_i}),\ VGG(\hat{P_i}^{GT})), \label{eq:12}
    \vspace{10pt}
\end{split}
\end{equation}
where the symbol $GT$ indicates the ground truth.


\section{Experiments}
We conduct extensive experiments on a publicly available benchmark dataset (\textit{i.e.}~ZJU-MoCap~\cite{peng2021neuralbody}) to verify the effectiveness of the proposed approach for human rendering under free-viewpoints with monocular videos.
We also illustrate the qualitative performance of YOTO in handling a larger number of individuals and in-the-wild settings with a single-time training on PeopleSnapshot~\cite{alldieck2018peoplesnapshot}.
Our findings indicate that YOTO can learn a considerable number of identities even from the in-the-wild videos, thus highlighting its effectiveness in handling such challenges.

\subsection{Datasets}
To thoroughly evaluate the performance of YOTO and to fairly compare against the baseline, we use ZJU-MoCap~\cite{peng2021neuralbody} dataset for quantitative evaluation.
As ZJU-MoCap dataset has 23 different camera views for each subject, we use camera \#1 for the training and the others for the evaluation.
We train a single copy of YOTO on all 6 different subjects (\textit{i.e.}, 377, 386, 387, 392, 393, 394) for the evaluation.
Therefore, all the qualitative and quantitative results on ZJU-MoCap in the following sections are rendered by one and the same YOTO, whereas 6 different HumanNeRF models are trained for 6 subjects separately.
For additional qualitative evaluation, we adopt PeopleSnapshot~\cite{alldieck2018peoplesnapshot} and use~22 different videos taken in various environments.

\begin{figure*}[!t]
\centering
  \includegraphics[width=1\linewidth]{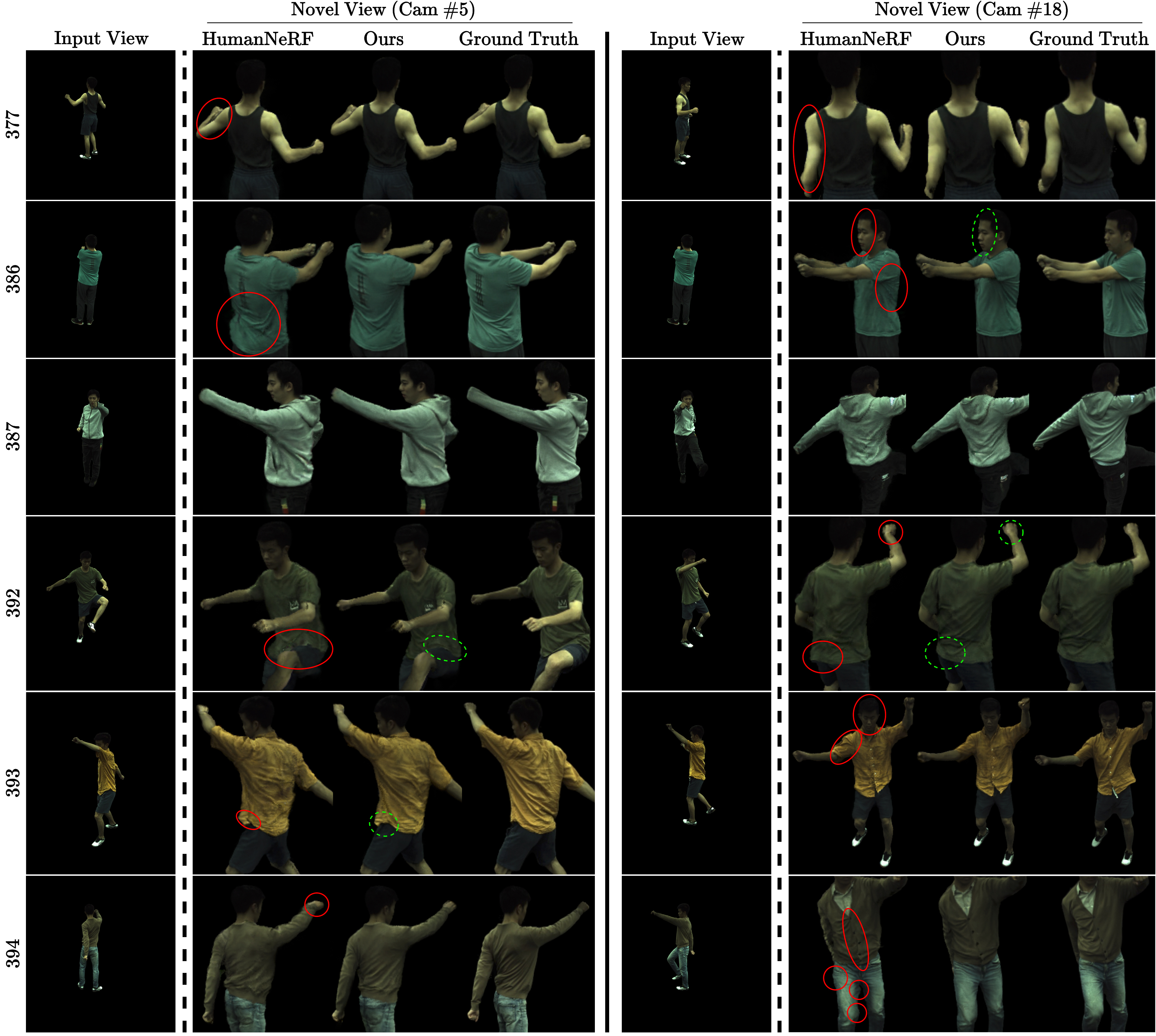}%
  \vspace{-7pt}
  \caption{Qualitative comparison of free-viewpoint synthesis against HumanNeRF~\cite{weng2022humannerf}. We indicate failing cases with red solid circles and successful cases with green dotted circles.}
  \vspace{-10pt}
  \label{fig:qualitative}
\end{figure*}
\begin{table}[t]
    \centering
    \resizebox{1\linewidth}{!}{
    \begin{tabular}{ l c c c }
        \toprule
        \multirow{ 2}{*}{Model} & Train Time & \# of Param. & Model Size \\
        & Hours & Million & MB \\
        \midrule
        HumanNeRF~\cite{weng2022humannerf} & 147  & 386.4  & 4428  \\
        YOTO (Ours) & \textbf{31}  & \textbf{65.3}  & \textbf{747}  \\
        \bottomrule
    \end{tabular}
    }
    \vspace{-6pt}
    \caption{Efficiency comparison between our proposed framework and  HumanNeRF~\cite{weng2022humannerf} in terms of training time, the total number of parameters, and the model size.}
    \label{tab:efficiency}
    \vspace{-15pt}
\end{table}
\subsection{Training Details}
We train our model with mostly the same configurations as HumanNeRF~\cite{weng2022humannerf} did so that we can clearly observe the benefits gained from our contributions.
We use Adam optimizer~\cite{kingma2015adam} with betas $(0.9, 0.999)$, and learning rates of $5\times10^{-4}$ for the canonical MLP $\mathbf{M}_{nr}$ and the subject codes $\mathbf{S}$, and $5\times10^{-5}$ for the others. The number of patches per iteration is $6$ with the dimension of $32\times32$ where each ray has 128 samples. We train all models including the baseline for 400K iterations with 1 Nvidia A100.

\begin{figure}[!t]
\begin{center}
   \includegraphics[width=\linewidth]{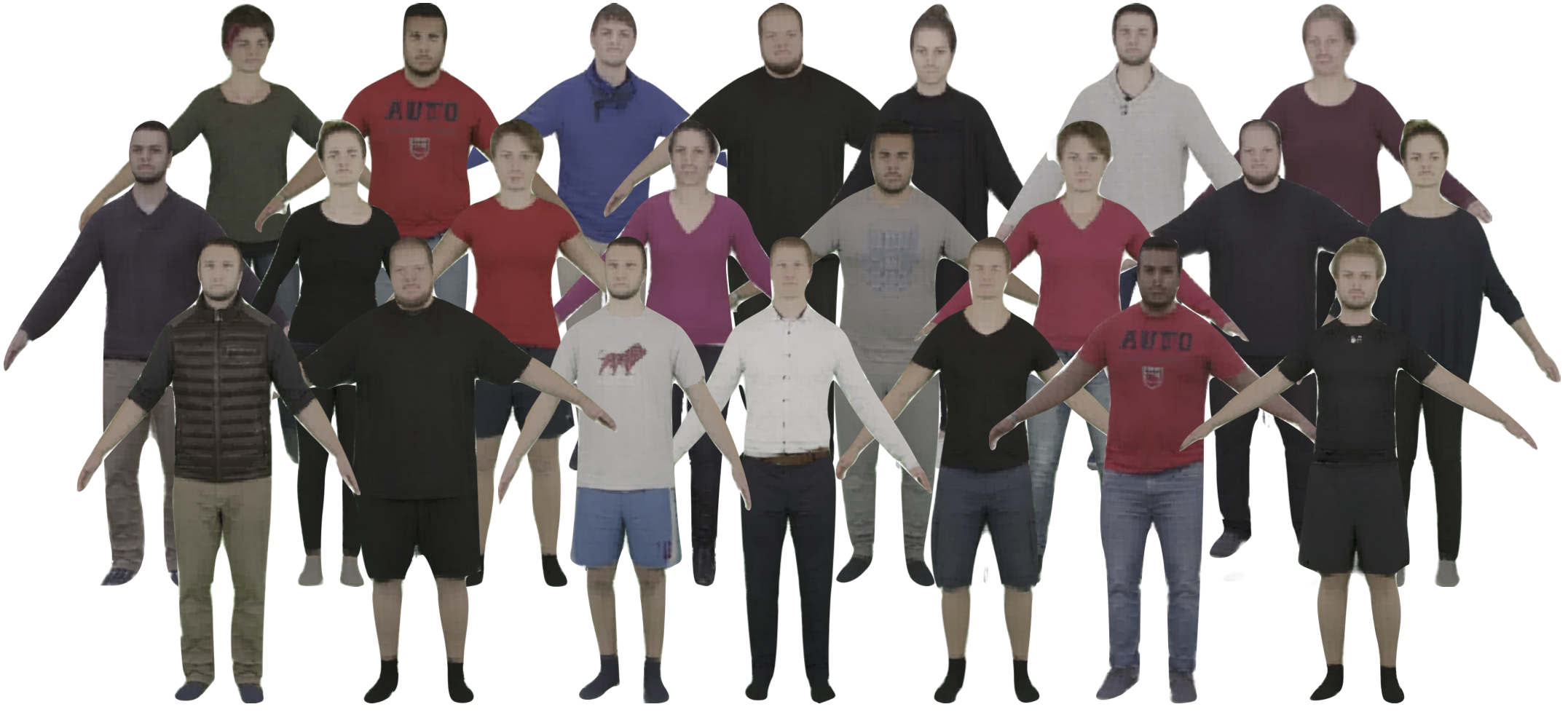}
   \vspace{-3.2em}
\end{center}
   \caption{Illustration of the capability of YOTO on PeopleSnapshot~\cite{alldieck2018peoplesnapshot} for learning with a larger number of input monocular videos. We show the rendering results of 22 different videos taken under various environments. Only one YOTO is used to learn all the representations.}
   \vspace{-7pt}
\label{fig:peoplesnapshot}
\end{figure}

\subsection{State-of-the-art Comparison}
Since HumanNeRF is the state-of-the-art method for free-viewpoint rendering with monocular video, we mainly compare our performance against it. 
For quantitative evaluation, Weng~\etal~\cite{weng2022humannerf} did not release their exact evaluation protocol. For instance, the exact frame IDs of the test set used for their evaluation are not available. Thus, we conduct the quantitative comparison by evaluating the different models using all the test frames, instead of sampling a subset from them. We believe this evaluation protocol is stricter to verify the performance of a model.
To ensure fair comparisons, we also directly utilize the pre-trained checkpoints released by Weng~\etal~\cite{weng2022humannerf} on all the following evaluations, as the authors confirmed that the best performances are from the released checkpoints. 

\par\noindent
\textbf{Quantitative comparison.} We compute PSNR, SSIM~\cite{wang2004ssim}, and LPIPS$^*$ and use these metrics to quantitatively evaluate our framework. As mentioned earlier, we use the released pre-trained checkpoints of HumanNeRF for the comparison and it is denoted as HumanNeRF$^*$, and evaluate the models on all the available novel-view frames instead of evaluating on sampled images. 
As demonstrated in Table~\ref{tab:quantitative}, YOTO achieves state-of-the-art performances in terms of all the metrics on ZJU-MoCap dataset. It can be observed that our framework shows considerable improvements on LPIPS$^*$ across all subjects, while PSNR and SSIM metrics are also clearly improved. It implies that YOTO renders more perceptually reasonable and coherent novel-view images compared to HumanNeRF~\cite{weng2022humannerf}. It should be noted that our YOTO framework jointly learns all the different identities via one-time training, while the results of HumanNeRF are evaluated on models separately trained on the different identities. These results can effectively demonstrate the performance advantages of our model.

\par\noindent
\textbf{Qualitative comparison.}\label{sec:qualitative}
Fig.~\ref{fig:qualitative} illustrates images rendered from novel views by both HumanNeRF and our proposed framework. The free-view rendering results for HumanNeRF are generated by their released checkpoints. As shown in Fig.~\ref{fig:qualitative}, HumanNeRF suffers from various artifacts caused by its motion field, non-rigid and canonical MLPs. The result of HumanNeRF on subject 377 shows that the unseen parts of arms are rendered as black, whereas our framework can coherently render the skin color. Moreover, the results also show that HumanNeRF suffers from incorrect pose transformation. We can observe that HumanNeRF fails to transform the head pose of subject 386, thus showing both eyes as indicated with a red circle. For subject 392, the baseline fails to model the non-rigid motions of the t-shirt's bottom hem caused by the raised left leg, while our framework can successfully render them benefiting from the proposed pose-conditioned identity codes for non-rigid motion estimation.
All other examples in Fig.~\ref{fig:qualitative} can further verify that YOTO is better at modeling coherent and pose-dependent non-rigid motions. 
\par In addition, we present qualitative results on PeopleSnapshot~\cite{alldieck2018peoplesnapshot} in Fig.~\ref{fig:peoplesnapshot}, to demonstrate the capability of our YOTO in joint handling a greater number of identities by one-time training.
As can be observed in Fig.~\ref{fig:peoplesnapshot}, mostly distinct appearances, especially the garments, are successfully learned by a single copy of YOTO. By controlling the capacity of MLPs of YOTO and the learning rate configurations based on the desired number of identities, we believe that the rendering quality can be further boosted. 

\begin{table}[t]
    \centering
    \small
    \begin{tabular}{ l | c c c }
        \toprule
        & PSNR\hspace{1pt}$\uparrow$ & SSIM\hspace{1pt}$\uparrow$ & LPIPS$^*$$\downarrow$ \\ \midrule 
{\footnotesize YOTO (Full Model)} & \textbf{30.51} & \textbf{0.9588} & \textbf{30.20} \\
{\footnotesize w/o pose-condition} & 30.41 & 0.9583 & 30.39 \\
\footnotesize w/o ID codes \& pose-condition & 28.72 & 0.9501 & 39.35 \\

        \bottomrule
    \end{tabular}
    \normalsize
    \vspace{-5pt}
    \caption{Quantitative model analysis on ZJU-MoCap for our novel contributions, \ie~the learnable identity codes and pose-conditioned code generation mechanism.}
    \label{tab:ablation}
    \vspace{-7pt}
\end{table}

\begin{figure}[!t]
\begin{center}
   \includegraphics[width=\linewidth]{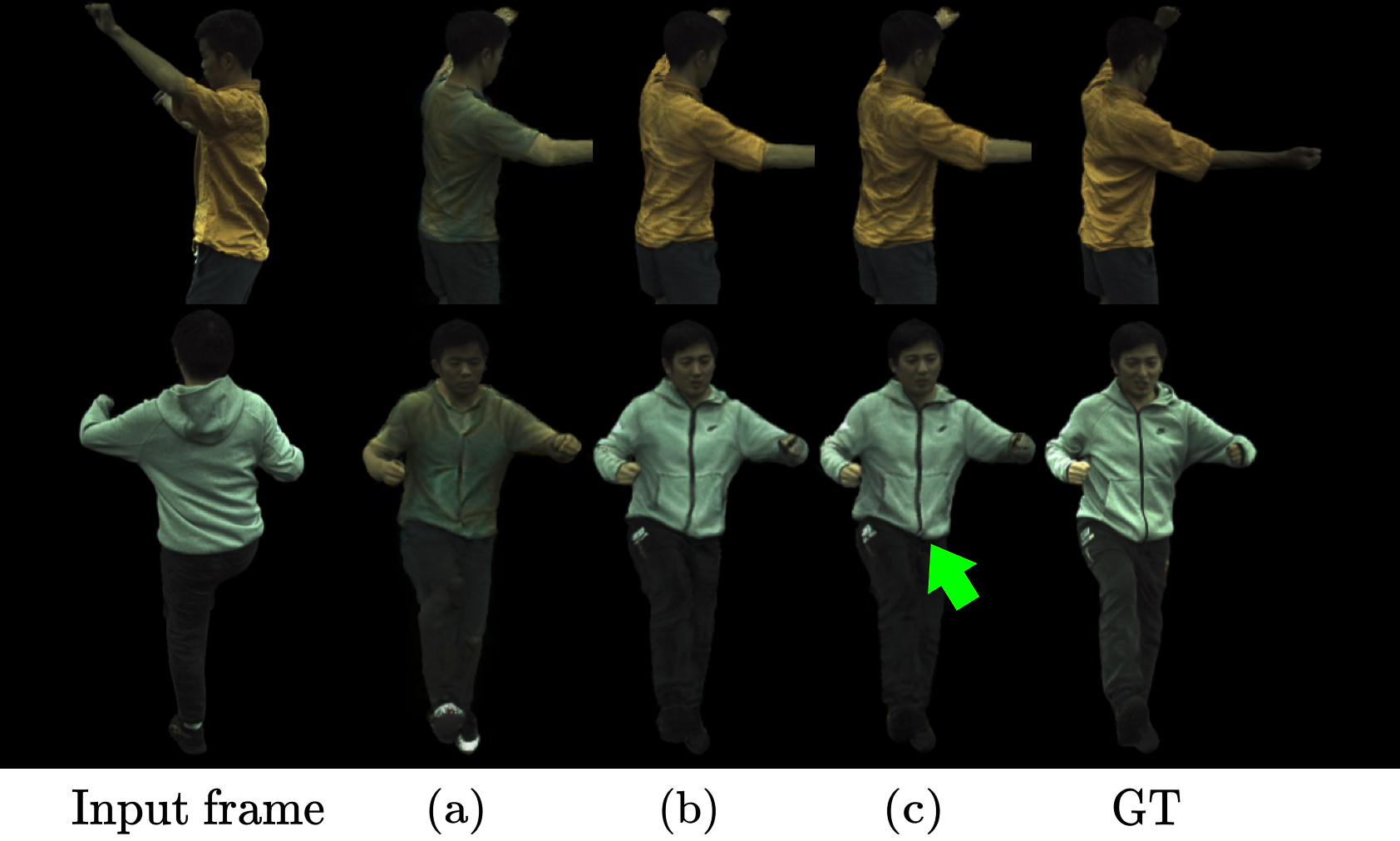}
   \vspace{-2.8em}
\end{center}
    \caption{Qualitative ablation on ZJU-MoCap. Without the identity codes and pose-conditioned non-rigid codes (a), the model fails to learn distinct appearances. The identity codes (b) resolve the issue of (a). The pose-conditioned non-rigid codes (c) allow the model to learn more coherent and high-fidelity non-rigid motions that correspond to the 3D pose.}
   \vspace{-7pt}
\label{fig:ablation}
\end{figure}

\par\noindent
\textbf{Efficiency comparison.} 
As shown in Table~\ref{tab:efficiency}, although we train our model with 6 different subjects jointly, there is no significant increase in the training time and the model size. YOTO requires approximately 5.5 seconds on average for every 20 iterations of training, while the baseline takes about 4.4 seconds. With the same computer resource available, YOTO boosts the total training time by $\times 4.7$ since HumanNeRF requires sequential training for all 6 subjects. Moreover, as also stated in Table~\ref{tab:efficiency}, the model size is only increased by 9MB which is 1.22\% of the original model size whereas it increases linearly with the number of subjects in the case of HumanNeRF.

\begin{figure}[!t]
\begin{center}
   \includegraphics[width=\linewidth]{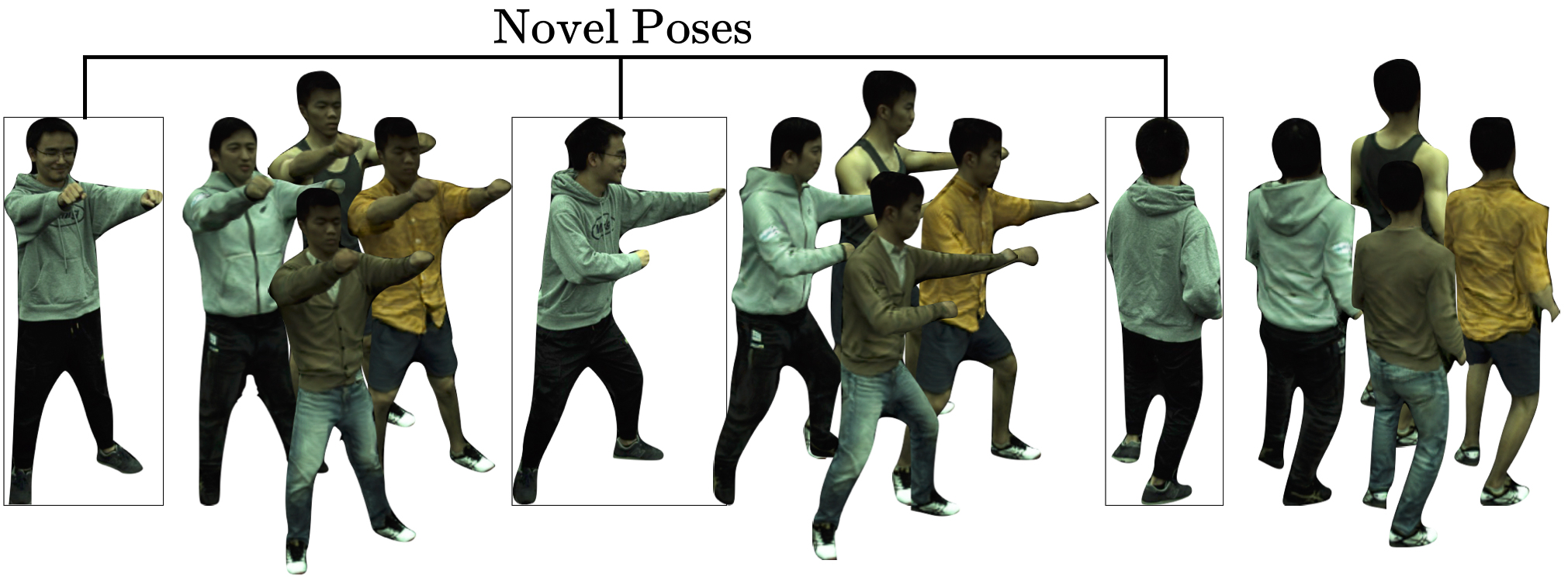}
   \vspace{-3.0em}
\end{center}
    \caption{Illustration of motion transfer on ZJU-MoCap.}
   \vspace{-15pt}
\label{fig:motion_transfer}
\end{figure}

\noindent
\textbf{Novel Motion Transfer.}
YOTO has an incidental advantage in that it can animate the learned identities by simply replacing the input pose with a novel one as illustrated in Fig.~\ref{fig:motion_transfer}. This improves the usability and applicability of YOTO since it would not need to train each model for the animation of each identity.

\section{Model Analysis}
To study the effectiveness of the proposed different components of YOTO, we consider different variants as shown in Table~\ref{tab:ablation}: (i) `YOTO (Full Model)' indicates the proposed full version of YOTO framework; (ii) `w/o pose-condition' denotes that we disable the pose-conditioned identity codes, while only using the learnable identity codes; (iii) `w/o ID codes \& pose-condition' means that we disable the learnable identity codes and the pose-conditioned codes. 
As shown in Table~\ref{tab:ablation}, the introduction of learnable identity codes significantly improves the quantitative performance, especially in terms of PSNR and LPIPS$^*$. The pose-conditioned code generation module further allows YOTO to achieve state-of-the-art performance.
Fig.~\ref{fig:ablation} also clearly demonstrates remarkable qualitative improvements made by both of our proposed contributions. For example, as indicated with a green arrow in Fig.~\ref{fig:ablation} the non-rigid motions of the hem of the jacket caused by the raised right leg is only coherently reproduced with the pose-conditioned codes.

\section{Conclusion}
In this paper, we presented a novel approach YOTO for simultaneous training of multi-identities from monocular videos for free-viewpoint rendering with higher fidelity and better efficiency.
By optimizing a single model with the proposed learnable identity codes, the model becomes able to handle all subjects in the input monocular videos, producing even higher quality compared to the models trained separately. We further propose a novel pose-conditioned identity code to enhance motion coherency in modeling. YOTO has fully demonstrated its effectiveness and established new state-of-the-art performance on the problem.

{\small
\bibliographystyle{ieee_fullname}
\bibliography{egbib}
}

\end{document}